\pdfoutput=1

\documentclass[11pt]{article}

\usepackage[]{acl}

\usepackage{times}
\usepackage{latexsym}

\usepackage[T1]{fontenc}

\usepackage[utf8]{inputenc}

\usepackage{microtype}
\usepackage{xcolor}
\usepackage{tabularx}
\usepackage{ragged2e}
\usepackage{pgfplots}
\pgfplotsset{compat=1.17}
\usepackage{CJKutf8}
\usepackage{caption,subcaption}
\usepackage{amsmath}
\usepackage{graphicx}
\usepackage{subcaption}
\usepackage{arydshln}
\usepackage{float}
\usepackage{booktabs}
\usepackage{soul}
\title{LiGCN: Label-interpretable Graph Convolutional Networks \\for Multi-label Text Classification}

\author{Irene Li$^1$, Aosong Feng$^1$, Hao Wu$^2$, Tianxiao Li$^1$,\\
\textbf{Toyotaro Suzumura}$^{3,4}$, \textbf{Ruihai Dong}$^5$\\
$^1$Yale University, USA, $^2$Trinity College Dublin, Ireland \\
$^3$Barcelona Supercomputing Center (BSC), Spain \\
$^4$University of Tokyo, Japan \\
$^5$University College Dublin, Ireland
}

\begin{document}
\maketitle
\begin{abstract}
Multi-label text classification (MLTC) is an attractive and challenging task in natural language processing (NLP). Compared with single-label text classification, MLTC has a wider range of applications in practice. In this paper, we propose a label-interpretable graph convolutional network model to solve the MLTC problem by modeling tokens and labels as nodes in a heterogeneous graph. In this way, we are able to take into account multiple relationships including token-level relationships. Besides, the model allows better interpretability for predicted labels as the token-label edges are exposed. We evaluate our method on four real-world datasets and it achieves competitive scores against selected baseline methods. Specifically, this model achieves a gain of 0.14 on the F1 score in the small label set MLTC, and 0.07 in the large label set scenario.  
\end{abstract}

\section{Introduction}
In the real world, we have seen an explosion of information on the internet, such as tweets, micro-blogs, articles, blog posts, etc. A practical issue is to assign classification labels to those instances. 
Such labels may be emotion tags for tweets and micro-blogs \cite{wang2016multi,li2020we}, or topic category tags for news, articles and blog posts \cite{yao2019graph}.  Multi-label text classification (MLTC) is the problem of assigning one or more labels to each instance. 

\begin{table}[t]
\small
\begin{tabular}{lp{4cm} p{1.5cm}} 
\toprule
& \textbf{Text}         & \textbf{Labels}        \\ \midrule[0.5pt]
S1& \begin{CJK*}{UTF8}{gbsn}
我不知道类似这样的\underline{困惑}到底还要持续多久。 
\end{CJK*} (I don’t know how long the \underline{confusion} like this will last.)                     & Anxiety       \\ \midrule[0.1pt]
S2 & nothing happened to \textbf{make me sad} but i almost \textbf{burst into tears} like 3 times today & Pessimism, Sadness\\ \midrule[0.1pt]
S3 & ...The price of BASF AG shares improved on Thursday due to its better than expected half year results. At 0900 GMT BASF was up 51 pfennigs at 42.75 marks...
&  C15, C151, C152, CCAT \\


\bottomrule           
\end{tabular}
\caption{Examples of multi-label emotion classification. Data source is explained in Sec.~\ref{sec:experiment}. Note that in S3, the labels are: C15 (Performance), C151 (Accounts/Earnings), C152 (Comment/Forecasts), CCAT (Corporate/Industrial). }
\label{tab:intro}
\vspace{-4mm}
\end{table}

Deep learning has been applied for MLTC due to their strong representation capacity in NLP tasks.
It has been shown that convolutional neural networks (CNNs) \cite{kim2014convolutional} achieve satisfying results for multi-label emotion classification \cite{wang2016multi,feng2018detecting}.
Besides, many recurrent neural networks (RNNs)-based models \cite{tang2015document} are also playing an important role \cite{huang2019seq2emo,yang2018sgm}. Recent breakthrough of pre-trained models, i.e., BERT \cite{devlin2019bert} and RoBERTa \cite{liu2019roberta}, achieved large performance gains in many NLP tasks. Existing work has applied BERT to solve MLTC problem successfully with very competitive performances \cite{li2019multi}.  Moreover, as a new type of neural network architecture with growing research interest, graph convolutional networks (GCNs) \cite{dblp:conf/iclr/kipfw17} have been applied to multiple NLP tasks. Different from CNN and RNN-based models, GCNs could capture the relations between words and texts if modeled as graphs \cite{yao2019graph,li2019should,li2020r}. In the paper, we focus on emphasising a GCN-based model to solve MLTC task. 

A major challenge for MLTC is the class imbalance. In practice, the number of labels may vary across the training data, and the frequency of each label may differ as well, bringing difficulties to model training \cite{quan2010blog}. In Table \ref{tab:intro}, we show some examples of tweet, micro-blog and news article, labeled with emotion tags or news topics.  As can be seen from those examples, there is a various number of coexisting labels. Another challenge is the interpretation of assigned class labels by figuring out the trigger words and phrases to corresponding labels. In the table, it is easy to tell that in S1, the emotion \textit{Anxiety} is very likely to be triggered by the word \textit{confusion}. However, S2 might be more complicated, with two possible triggering phrases \textit{makes me sad} and \textit{burst into tears} and two emotion labels. There might be different opinions on which phrase triggers which emotion.

To tackle the mentioned challenges and investigate different perspectives, we propose label-interpretable graph convolutional networks for MLTC. We model each token and class label as nodes in a heterogeneous graph, considering various types of edges: token-token, token-label, and label-label. Then we apply graph convolution to graph-level classification. As GCN works well in semi-supervised learning \cite{DBLP:conf/asunam/GhorbaniBR19}, we can then ease the impact of data imbalance.  Finally, since the token-label relationships are exposed in the graph, one can easily identify the triggering tokens to a specific class, providing a good interpretability for multi-label classification.

The \textbf{contributions} of our work are as follows: (1) We transfer the MLTC task to a link prediction task within a constructed graph to predict output labels. In this way, our model is able to provide token-level interpretation for classification.  (2) To the best of our knowledge, this is the first work that considers token-label relationships within a manner of a graph neural network for MLTC, allowing label interpretability.
(3) We conduct extensive experiments on four representative datasets and achieve competitive results. We also demonstrate comprehensive analysis and ablation studies to show the effectiveness of our proposed model for label nodes and token-label edges. We release our code in \url{https://github.com/IreneZihuiLi/LiGCN}.

\section{Related Work}

\textbf{Multi-label Text Classification}
Many existing works focus on single-label text classification, while limited literature is available for multi-label text classification. In general, these methods fall into three categories: problem transformation, label adaptation and transfer learning. Problem transformation is to transform the muli-label classification task into a set of single-label tasks \cite{jabreel2019deep,zhang2020topic}, but this method is not scalable when the label set is large. 
Label adaptation is to rank the predicted classes or set a threshold to filter the candidate classes. \citet{chen2017ensemble} proposed a novel method to apply an RNN for multi-label generation with the help of text features learned using CNNs. Transfer learning focuses on utilizing knowledge learned to unknown entries. \citet{xiao2021head} proposed a model which transfers the meta-knowledge from data-rich labels to data-poor labels. 
Moreover, some models also take label correlations into consideration, such as Seq2Emo \cite{huang2019seq2emo} and EmoGraph \cite{xu2020emograph}. However, some of them may ignore the relationships between input tokens and class labels, making them less interpretable. 
Please note that there is a research topic named extreme multi-label text classification \cite{liu2017deep}, where the pool of candidate labels is extremely large. However, we do not target on the extreme case.

\textbf{Graph Neural Networks in NLP} 
Previous research has introduced GCN-based methods for NLP tasks by formulating them as graph-structured data tasks. 
A fundamental task is text classification. Many works show that it is possible to utilize inter-relations of documents or tokens to infer the labels \cite{yao2019graph,zhang2019aspect}. Besides, some NLP tasks focus on learning relationships between nodes in a graph, such as concept prerequisites \cite{li2019should} and leveraging dependency trees predicted by GCNs for machine translation \cite{bastings-etal-2017-graph}. 
Recently, variations of GCN models have been investigated for general text classification tasks \cite{linmei2019heterogeneous,tayal2019short,ragesh2021hetegcn}. Limited efforts have been made to apply GCNs for multi-label text classification. 
For example, EmoGraph \cite{xu2020emograph} is a model that captures the dependencies among emotions through graph networks. 


\section{Method}

In this section, we first provide task definition and preliminary, then we introduce the proposed model for multi-label text classification. 

\subsection{Task Definition}
In multi-label text classification task, we are given the training data $\{D,Y\}$ . For the $i$-th sample,  $D^i$ contains a list of tokens $D^i=\{w_1,w_2,...w_m\}$ and $Y^i$ is a list of binary labels $Y^i=\{y_1,y_2,...y_n\}$, $y$ is 1 if the class label is positive, 0 otherwise. The size of label set $n$ can be small or large.
In testing, we predict labels $\hat{Y}_{test}^i$ given $D_{test}^i$ .

\subsection{Preliminary}
Graph convolutional network (GCN) \cite{dblp:conf/iclr/kipfw17} is a type of deep architecture for graph-structured data. In a typical GCN model, we define a graph as $G=(\mathcal{V},\mathcal{E})$, where $\mathcal{V}$ is a set of nodes and $\mathcal{E}$ is a set of edges.  Normally, the edges are represented as an adjacency matrix $A$, and the node representation is defined as $X$. In a multi-layer GCN, the propagation rule for layer $l$ is defined as:
\begin{align} 
    H^{(l)} =\sigma\left(norm(A^{(l-1)})H^{(l-1)} W^{(l-1)}\right),  \label{eq:gcn} 
\end{align} 
where $norm(A) = \tilde{D}^{-\frac{1}{2}} \tilde{A} \tilde{D}^{-\frac{1}{2}}$ is a normalization function, $H$ denotes the node representation, and $W$ is the parameter matrix to be learned. $\tilde{A} = A+I_{|\mathcal{V}|}$, $\tilde{D}$ denotes the degree matrix of $\tilde{A}$, In general, in the very first layer, we have $H^{(0)}=X$.


\subsection{Label-interpretable Graph Convolutional Networks}

In this paper, we propose the LiGCN model, which allows interpretation on the labels when doing MLTC.
For each training sample, we construct an undirected graph. We define two types of nodes: token node and label node, and the node representations are $X_{token}$ and $X_{label}$.  Therefore, there are three types of relations between the nodes, defined by the adjacency matrices: $A_{token}$ (between token nodes), $A_{label}$ (between label nodes) and $A_{token\_label}$ (between token nodes and label nodes). 



\begin{figure*}[t]
    \centering
    \includegraphics[width=\linewidth]{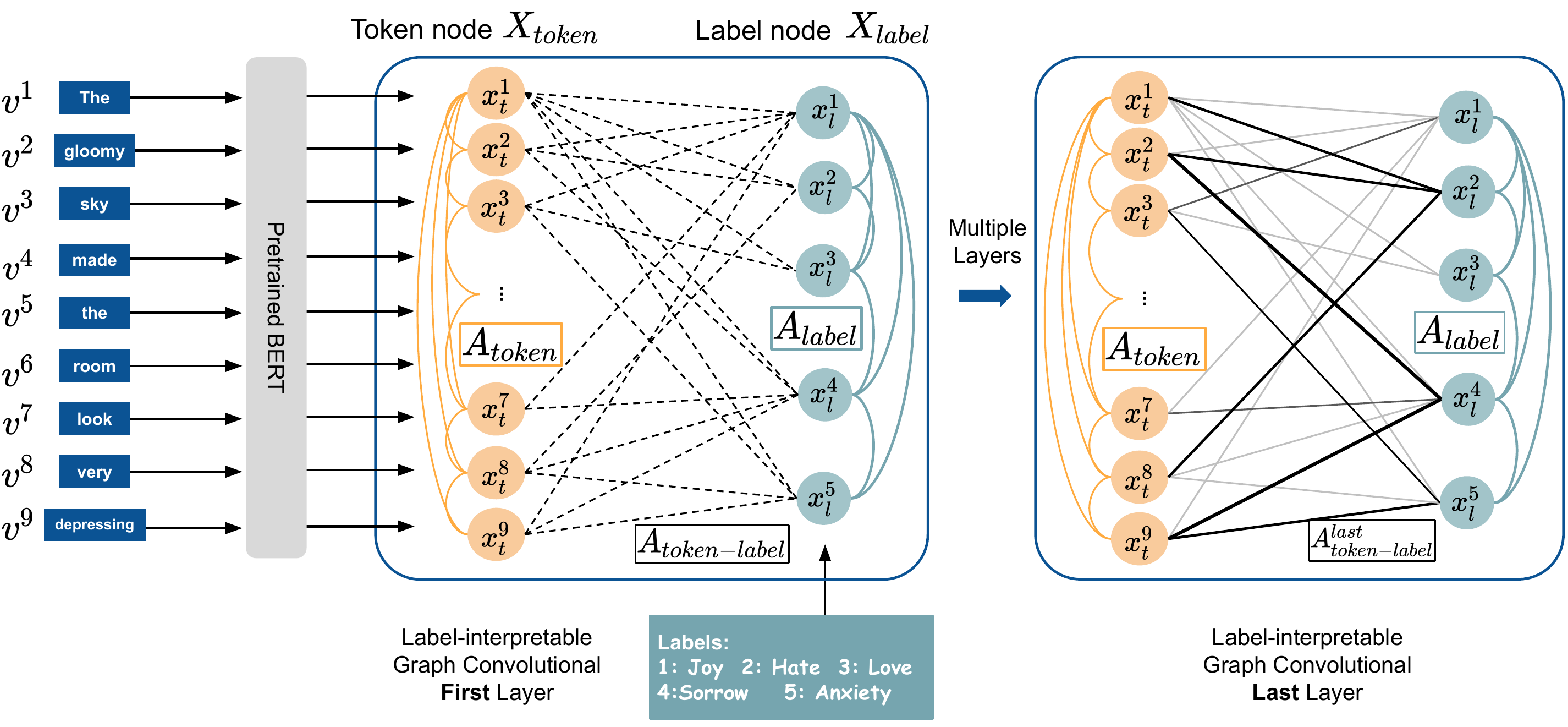}
  \caption{LiGCN model overview. (Best viewed in color.)}
\label{fig:model}
\end{figure*}

We show the model overview in Figure \ref{fig:model}. It consists of two main components: a pre-trained BERT/RoBERTa encoder\footnote{https://huggingface.co/bert-base-multilingual-cased,\\https://huggingface.co/roberta-base} and label-node graph convolutional layers. In the LiGCN model, we have a list of token nodes  $X_{token}$ in orange ellipses, and a list of label nodes  $X_{label}$ in blue ellipses. Besides, there are edges between token nodes $A_{token}$, edges between label nodes $A_{label}$, and edges between token and label nodes $A_{token\_label}$. We explain them in greater details below.




\textbf{Node Representations $\boldsymbol X$}:
In the very first layer, to initialize token nodes, we encode the input data $D^i=\{w_1,w_2,...w_m\}$ using a pre-trained BERT or RoBERTa model and possible other BERT-based ones, where we take the representation of each token including  \texttt{[CLS]} token as $X_{token}$. For the label nodes, we initialize them using one-hot vectors.

\textbf{Adjacency Matrix $\boldsymbol A$}:
In our experiments, to initialize token node adjacency matrix $A_{token}$, we use the token nodes to construct an undirected chain graph, where we consider an input sequence as its natural order, i.e., in $A_{token}$: $A_{i,i+1}=1$. Since it is an undirected graph, the adjacency matrix is symmetric, i.e., $A_{i+1,i}=1$. We also add self-loop to each token: $A_{i,i}=1$.  In other words, $A_{token}$ is a symmetric $m$-by-$m$ matrix with an upper bandwidth of 1, where $m$ is the number of token nodes. 

We initialize $A_{label}$ with an identity matrix, and $A_{token\_label}$ with a zero matrix. In the later layers, we reconstruct $A_{token\_label}$ for layer $l$ by applying cosine similarity between $X_{token}$ and $X_{label}$ of the current layer:
\begin{equation}
    A^{l}_{token\_label} = cosine(X^{l}_{token},X^{l}_{label}).
    \label{eq:reconstruct}
\end{equation}

The value is normalized into the range of [0,1]. After this update, the model conducts graph convolution operation as in Eq.~\ref{eq:gcn}. 

In Figure \ref{fig:model}, we are not showing self-loops, so $A_{label}$ is not visible. We show only a subset of edges from $A_{token}$ and $A_{token\_label}$. Note that we use dashed lines at the first LiGCN layer because $A_{token\_label}$ is a zero matrix. 

We also investigate other possible ways to build $A_{token}$ including dependency parsing trees \cite{huang2020syntax} and random initialization, but our method gives the best result. Such ways may not bring useful information to the graph: the help from dependency relations may be limited in the case of classification, and random initialization brings noises. As we focus more on the network convolution, we leave investigating more methods for initialization as future work.

\begin{table*}[t]
\small
\centering
\begin{tabular}{lrrrrrrrr}
\toprule
\textbf{Dataset} & \textbf{\#train} & \textbf{\#dev} & \textbf{\#test} & \textbf{\#class} & \textbf{\#avg label}  &\textbf{\#token max}& \textbf{\#token median} &\textbf{\#token mean}  \\ \midrule[0.5pt]
\textbf{SemEval}  & 6,839 &  887 &  3,260  &        11   & 2.37         &  499     & 26  & 32.08\\
\textbf{RenCECps}   & 27,299 &  -  &  7,739  &      8         &  1.37  & 36     & 17    & 16.42 \\
\textbf{RCV1}      & 20,647 &   3,000   &  783,144        &         103        &  3.20  &  9,380  & 198 & 259.06\\
\textbf{AAPD}      & 54,840 &  -   &   1,000      &         54        & 2.41   &  500  & 157 & 166.41\\
\bottomrule 
\end{tabular}
\caption{Dataset statistics on four selected corpora.}
\label{tab:data}

\end{table*}

\textbf{Predictions} In the last LiGCN layer, we are able to reconstruct $A^{last}_{token\_label}$ using Eq.~\ref{eq:reconstruct}. For each label node $j$, we sum up the edge weights from $A^{last}_{token\_label}$ to get a score, 

\begin{equation}
    score(j) = \displaystyle\sum_{v_i\in \mathcal{V}_{token}} A^{last}_{i,j},
    \label{eq:pred}
\end{equation}
 where $\mathcal{V}_{token}$ is the set of all token nodes in the last LiGCN layer. Then we apply a softmax function over all the labels, so that the scores are transformed to probabilities of labels. Finally, to make the prediction, we rank the probabilities in a descending order, and keep the top $k$ labels from the ranking as predictions.
As the predictions are in forms of probabilities, we also convert the ground truths into probability distribution. We use the mean square error (MSE) as the loss function. Another way is to apply the normal cross-entropy for classification, but it achieves slightly worse results, so we do not include it.

\section{Experimental Results}
\label{sec:experiment}

We evaluate on four public datasets, summarized in Table \ref{tab:data} and \ref{tab:labeldata}: \textbf{SemEval}
\cite{mohammad2018semeval} contains a list of subtasks on labeled tweets data. In our experiments, we focus on Task1 (E-c) challenge on English corpus: multi-label classification tweets on 11 emotions. 
\textbf{RenCECps} \cite{quan2010blog} is a Chinese blog corpus which contains manual annotation of eight emotional categories.
It not only provides sentence-level emotion annotations, but also contains word-level annotations, where in each sentence, emotional words are highlighted. 
\textbf{RCV1} \cite{lewis2004rcv1} consists of manually-labeled English news articles from Reuters Ltd. Each news article has a list of topic class labels, i.e., CCAT for Corporate/industrial, G12 for Internal politics. We follow the same setting of \citet{yang2018sgm} and \citet{nam2017maximizing}, and do MLTC on the top 103 classes.  
\textbf{AAPD} \cite{yang2018sgm} is a set of English computer science paper abstracts and corresponding subjects from \url{arxiv.org}.

\begin{table}[t]
\centering
\footnotesize
\begin{tabular}{crrrr}
\toprule
\textbf{\#label}           & \textbf{SemEval} & \textbf{RenCEPcs} & \textbf{RCV1} & \textbf{AAPD} \\\midrule[0.5pt]
0  & 293     & 2,755     & 0    & 0  \\
1  & 1,481    & \textbf{18,858}    & 35,591   & 0 \\
2 & \textbf{4,491}    & 11,417    & 203,030   & \textbf{38,763}\\
3 & 3,459    & 1,815     & \textbf{362,124}   & 12,782\\
4 & 1,073    & 172      & 85,527    & 3,229\\
$\geq 5$        & 186     & 21       & 120,518  & 1,066\\ \midrule[0.1pt]
Avg.   &2.37   &1.37   &3.20 & 2.41 \\
\bottomrule 
\end{tabular}
\caption{Label number distributions.}
\label{tab:labeldata}
\vspace{-3mm}
\end{table}

We report the following evaluation metrics: 

\textbf{Micro/Macro F1, Jaccard Index} We report micro-average and macro-average F1 scores as did by previous works \cite{baziotis2018ntua,huang2019seq2emo} if the label set is small. 
Besides, we follow the Jaccard index used by \cite{mohammad2018semeval,baziotis2018ntua,huang2019seq2emo}, as always referred as multi-label accuracy. The definition is given below:
\vspace{-2mm}
\begin{equation}
    \nonumber
    J=\frac{1}{N} \sum_{i=1}^{N} \frac{\left|Y^{i} \cap \hat{Y}^{i}\right|}{\left|Y^{i} \cup \hat{Y}^{i}\right|},
    \vspace{-2mm}
\end{equation}
where $N$ is the number of samples, $Y^{i}$ denotes the ground truth labels and $\hat{Y}^{i}$ denotes system predicted labels. 

\textbf{P@k and nDCG@k} When the label set is large, we also report widely-applied metrics P@K and nDCG@K ($k=1,3,5$). 

We apply two graph convolutional layers for all datasets by default for our LiGCN model. In Table \ref{tab:para}, we show the hyper-parameters conducted in our experiments. We use 4.00E-06 as the learning rate for all experiments. Since we use two LiGCN layers, \texttt{hid dim1} is the first layer hidden dimension number, and \texttt{hid dim2} is the second layer hidden dimension number. These hyper-parameters were selected by dev sets (if exist), otherwise selected by manual tuning with about 5-10 rounds for search trials.

\begin{table}[t]
\centering
\footnotesize
\begin{tabular}{lrrrr} 
\toprule
                     & \textbf{SemEval}  & \textbf{RenCECPs} & \textbf{RCV1}  & \textbf{AAPD}  \\  \midrule[0.5pt]
seq length      & 17       & 32        & 256    & 256 \\
hid dim1          & 64       & 64        & 256    & 256   \\
hid dim2          & 16       & 16        & 64    & 64   \\
epoch num          & 5        & 3         & 10       & 10   \\ 
top-$k$               & 2        & 1         & 5     & 5   \\
\bottomrule 
\end{tabular}
\caption{Hyper-parameters chosen in our experiments. }
\label{tab:para}
\end{table}


\begin{table*}[th]
\centering
\small
\begin{tabular}{llllclll}\toprule
  &  \multicolumn{3}{c}{SemEval} & & \multicolumn{3}{c}{RenCECps} \\ \cline{2-4} \cline{6-8} 
 \textbf{Method}          & \textbf{Macro F1}    & \textbf{Micro F1}     & \textbf{Jaccard} &   & \textbf{Macro F1}    & \textbf{Micro F1}     & \textbf{Jaccard}  \\ \midrule 
SGM \cite{yang2018sgm} & 0.4110 & 0.5750 & 0.4820 && - & 0.5560 & -\\
Seq2Emo \cite{huang2019seq2emo}     &-    & 0.7089  & 0.5919  && - & - & - \\
TECap \cite{zhang2020topic} &0.5760 & 0.6820  & - && 0.4550   & 0.5310   & -  \\
MEDA \cite{deng2020multi} &  - & - & - && 0.4831 & 0.6076 & - \\
EmoGraph (BERT-GCN)*\cite{xu2020emograph} & \underline{0.6367}     & \underline{0.8108}   & \underline{0.6818} &&  \underline{0.6129}   & \underline{0.8559}   & \underline{0.7481}  
\\
\midrule[0.1pt]
BERT      & 0.5223        & 0.6454   & 0.4766 &&  0.5344   & 0.6365   & 0.4669 \\
RoBERTa     &    0.5039     & 0.6817   & 0.5171  && 0.5842   & 0.7987   & 0.6649 \\
BERT-LiGCN (\textbf{ours})  & 0.7368     & 0.8312   & 0.7111  && 0.7138   & 0.8615   & 0.7567 \\
RoBERTa-LiGCN (\textbf{ours})  &   \textbf{0.7786}      & \textbf{0.8579}   & \textbf{0.7512}  && \textbf{0.7429}   & \textbf{0.8756}   & \textbf{0.7787} \\ 
\bottomrule
\end{tabular}
\caption{Evaluation results on SemEval-2018 and RenCECps. \textbf{BERT/RoBERTa} means the system which has a linear layer on top of the original BERT/RoBERTa. EmoGraph*:we present results of our own optimized implementation of EmoGraph. Underlined scores are the best ones among baselines.}
\label{tab:semevalren}
\end{table*}

\begin{table*}[t]
\centering
\small
\begin{tabular}{lcccccc}
\toprule
\textbf{Method} & \textbf{P@1} & \textbf{P@3} & \textbf{P@5} & \textbf{nDCG@3} & \textbf{nDCG@5} & \textbf{F1-score} \\ \midrule[0.5pt]
\textbf{\textit{RCV1}}            &              &              &              &                 &                 &                   \\
XML-CNN \cite{liu2017deep}         & 95.75        & 78.63        & 54.94        & \underline{89.89}           & 90.77           & 75.92             \\

Imprinting   \cite{qi2018low}    & 77.38        & 47.96        & 31.45        & 58.83           & 57.91           & 26.35             \\
DXML  \cite{zhang2018extreme}          & 94.04        & 78.65        & 54.38        & 89.83           & 90.21           & 75.76             \\
OLTR \cite{liu2019large}            & 93.79        & 61.36        & 44.78        & 74.37           & 77.05           & 56.44             \\

BBN  \cite{zhou2020bbn}           & 94.61        & 77.98        & 54.25        & 88.97           & 89.68           & 78.65             \\
HTTN  \cite{xiao2021head}          & \underline{95.86}        & \underline{78.92}        & \underline{55.27}        & 89.61           & \underline{90.86}           & \underline{77.72}             \\
BERT-LiGCN (\textbf{ours})     & 94.42        & 80.98        & 55.48       & 91.93           & 91.94           & 83.14             \\
RoBERTa-LiGCN (\textbf{ours})  & 95.61        & \textbf{82.40}        & \textbf{56.31}        & \textbf{93.40}           & \textbf{93.26}           & \textbf{83.66}             \\
                \midrule[0.5pt]
\textbf{\textit{AAPD}}            &              &              &              &                 &                 &                   \\
XML-CNN \cite{liu2017deep}        & 74.38        & 53.84        & 37.79        & 71.12           & 75.93           & 65.35             \\
Imprinting \cite{qi2018low}     & 68.68        & 38.22        & 23.71        & 55.30           & 55.67           & 25.58             \\
DXML \cite{zhang2018extreme}           & 80.54        & 56.30        & 39.16        & 77.23           & 80.99           & 65.13             \\

OLTR \cite{liu2019large}           & 78.96        & 56.28        & 38.60        & 74.66           & 78.58           & 62.48             \\

BBN   \cite{zhou2020bbn}           & 81.56        & 57.81        & 39.10        & 76.92           & 80.06           & 66.73             \\
HTTN \cite{xiao2021head}           & \underline{83.84}        & \underline{59.92}        & \underline{40.79}        & \underline{79.27}           & \underline{82.67}           & \underline{69.25}             \\
BERT-LiGCN (\textbf{ours})      & \textbf{84.10}        & \textbf{61.33}        & 40.88        & \textbf{80.77}           & 83.68           & 75.89             \\
RoBERTa-LiGCN (\textbf{ours})   & 82.50        & 61.26        & \textbf{41.38}        & 80.39           & \textbf{83.83}           & \textbf{76.25}            \\
\bottomrule 
\end{tabular}
\caption{Results on RCV1 and AAPD: note that BERT/RoBERETa baseline have a negative result in this case. }
\label{tab:res_large}

\end{table*}

\subsection{Small Label Sets}

We first evaluate the proposed model on SemEval and RenCECps in Table \ref{tab:semevalren}. Both of them have a small label set, so we report Macro, Micro F1 and Jaccard. We select the following baselines: SGM \cite{yang2018sgm} applies a sequence generation model and a decoder structure; 
Seq2Emo \cite{huang2019seq2emo} is an LSTM-based model that takes into account the correlations among target labels;
TECap \cite{zhang2020topic} is a topic-enhanced capsule network, which contains a variational autoencoder and a capsule module for multi-label emotion detection; MEDA \cite{deng2020multi} is a multi-label emotion detection architecture that focuses on detecting all emotions shown in the text, and it takes BERT for sentence encoding. Finally, EmoGraph \cite{xu2020emograph} is a graph-based method that learns dependencies among emotion nodes using GCNs. The result presented is based on our implementation with optimized parameters, and is slightly better than their original paper. We also compare with a BERT and a RoBERTa model as baselines (BERT, RoBERTa). We first take the representation of \texttt{[CLS]} token from pre-trained BERT/RoBERTa, on top of that, a linear layer is connected. For the two datasets, we set the top-$k$ value to be the average number of labels in each dataset.

We can observe that our model surpasses all the selected baselines in most of the cases. Especially, both MEDA and EmoGraph applied pre-trained BERT model as our BERT-LiGCN model does, and we significantly outperform those models on all three metrics. Moreover, EmoGraph is a similar model with LiGCN but it only considers a single node type (class node) while LiGCN considers both class node and token node. This shows that, with a much complex graph structure, LiGCN is able to capture more information when doing classification. Besides, LiGCN benefits from using RoBERTa as the encoder, as RoBERTa improves upon BERT by a small margin. 

\begin{table*}[t]
\small
\centering
\begin{tabular}{llllclll}\toprule
  &  \multicolumn{3}{c}{SemEval} & & \multicolumn{3}{c}{RenCECps} \\ \cline{2-4} \cline{6-8} 
 \textbf{BERT-LiGCN}          & \textbf{Macro F1}    & \textbf{Micro F1}     & \textbf{Jaccard} &   & \textbf{Macro F1}    & \textbf{Micro F1}     & \textbf{Jaccard}  \\ \midrule[0.5pt]
1-layer & 0.7131 & 0.8159 & 0.6891 && 0.7054 & 0.8091 & 0.6794\\
2-layer  & \textbf{0.7368}     & \textbf{0.8312}   & \textbf{0.7111}  && \textbf{0.7138}   & \textbf{0.8615}   & \textbf{0.7567} \\
3-layer    & 0.7109   & 0.8145  & 0.6871 && 0.7044 & 0.8085 & 0.6785  \\
\bottomrule 
\end{tabular}
\caption{Ablation study on number of layers: compare numbers of BERT-LiGCN layer.}
\label{tab:ablation}
\end{table*}

\subsection{Large Label Sets}
We then evaluate large label sets using RCV1 and AAPD, shown in Table \ref{tab:res_large}. We compare with a number of recent baselines: XML-CNN \cite{liu2017deep} applied a CNN and dynamic pooling to learn features for MLTC; Imprinting \cite{qi2018low} is a weight imprinting method that directly set the final layer weights of deep models from new training examples; DXML \cite{zhang2018extreme} focused on label co-occurrence graph to solve the multi-label long-tail issue;  OLTR \cite{liu2019large} is a method that handles long-tail and imbalanced classification problems; BBN \cite{zhou2020bbn} is a model that considers both representation learning and classifier learning; HTTN \cite{xiao2021head} learns the meta-knowledge so as to transfer from data-rich head labels to data-poor ones. We set the top-$k$ value to be $k=5$ in the prediction so as to evaluate P@$k$ and nDCG@$k$. 

Our model can also surpass the selected baselines in general, especially with a large improvement on the F1-score for the two datasets. Surprisingly,  BERT-LiGCN performs better than RoBERTa-LiGCN on P@1, P@3 and nDCG@3 in AAPD. In other words, BERT-LiGCN can predict better top-3 candidates, while RoBERTa-LiGCN can do well in predicting the 4-th and 5-th candidates. But in both BERT and RoBERTa settings, our model can perform close and better compared with these recent baselines. 


\section{Analysis}

In this section, we first focus on ablation study of the proposed model. We then demonstrate the interpretability for labels with several case studies where we identify keywords that trigger certain labels.
Moreover, we study and examine the meaning of label representations learned by the model.

\subsection{Ablation Study}



We first study the impact of graph convolutional layer numbers in Table \ref{tab:ablation}. We test with 1, 2 and 3 LiGCN layers on SemEval and RenCECps using BERT-LiGCN model. In general, we see that 2-layer is the best setting. Less layers may not be enough for information exchange within nodes in the GCN models. While increasing the layer number results in training difficulties and lower performances, as some other works \cite{li2018deeper} have shown.


\subsection{Token-label Relations}

\begin{figure}[t]
    \centering 

\begin{subfigure}{0.48\textwidth}
  \includegraphics[width=\linewidth]{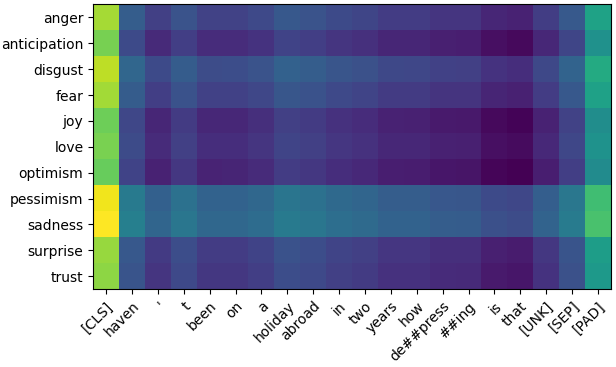}
  \caption{An example from SemEval.
  }
  \label{fig:exp1}
\end{subfigure}\hfil 


\begin{subfigure}{0.5\textwidth}
  \includegraphics[width=\linewidth]{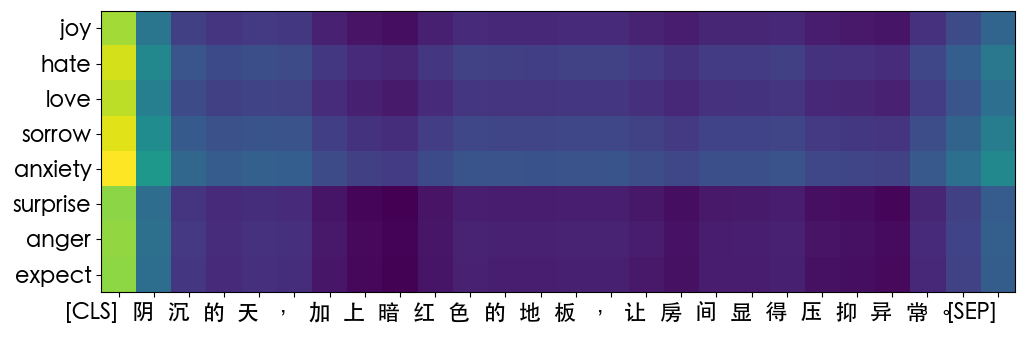}
  \caption{An example from RenCECps.}
  \label{fig:exp3}
\end{subfigure}

\caption{Visualization of word-label weights: brighter color indicates a larger value.}
\label{fig:case_study}
\end{figure}

As our proposed model provides token-label relations, we further study token-level explanations via case studies and a quantitative analysis. 

\textbf{Case Study} 
We show examples by visualizing the token-label weights in Figure \ref{fig:case_study}. Specifically,
we take the reconstructed $A_{token\_label}$ and normalize the matrix so that all values sum up to 1. 
We select a sample from SemEval test set, as shown in Figure \ref{fig:exp1}: \textit{haven't been on a holiday abroad in two years how depressing is that... } (labels: \texttt{pessimism} and \texttt{sadness}). 
In such a heatmap, columns are tokens while rows are the emotion labels. 
We can notice that our model computes a higher score to the text chunk \textit{haven't} and \textit{holiday abroad}, and a relatively lower score to \textit{depressing}, by looking at the corresponding columns. Therefore, the prediction being \texttt{pessimism} and \texttt{sadness} is mostly triggered by \textit{haven't} and \textit{holiday abroad}. This indicates that the emotion label to be such a negative sentiment is because this person \lq haven't been on a holiday abroad.\rq   Even though there is a strong negative sentiment word \textit{depressing}, LiGCN attempts to pick out implicit and deeper reasons. Besides, one can see that our model highlights other three close emotions \textit{anger}, \textit{disgust} and \textit{fear}. 
\footnote{When doing classification, both special tokens of BERT \texttt{[CLS]} and \texttt{[SEP]} contain useful semantic information of the whole sequence, so the color tends to be brighter.}

We show another example from RenCECps test set in Figure \ref{fig:exp3}: \begin{CJK*}{UTF8}{gbsn}\underline{阴沉}的天，加上暗红色的地板，让房间显得\underline{压抑}异常。\end{CJK*} (\textit{The \underline{gloomy} sky, together with the dark red floor, made the room look very \underline{depressing}.}) The ground truth labels are \texttt{Sorrow} and \texttt{Anxiety}. Our model successfully predicts these two class labels; moreover, the model also suggests that \texttt{Hate} is a possible label, which is reasonable in this particular example. 
Besides, in the annotation of the original dataset by \cite{quan2010blog}, we find that two keywords are highlighted for this example: \begin{CJK*}{UTF8}{gbsn}阴沉\end{CJK*} (gloomy) : Surprise=0, Sorrow=0, Love=0, Joy=0, Hate=0,  Expect=0,  Anxiety=0.6, Anger=0; and \begin{CJK*}{UTF8}{gbsn}压抑\end{CJK*} (depressing):  Surprise=0, Sorrow=0.5, Love=0, Joy=0, Hate=0, Expect=0,  Anxiety=0.7, Anger=0. Our model also captures such a trend successfully by showing a higher score near or on these token columns.

\textbf{Quantitative Analysis} So far, we have demonstrated that our model is able to identify the triggering words for each individual class from the confidence score of the token-label edges. To quantitatively show the quality of identified triggering words, we compute MSE between our best performed model and the ground truth annotations for the test set of RenCECps. 
Similar to previous analysis, we first normalize the constructed token-label adjacency matrix $A_{token\_label}$, then construct a token-label matrix $A_{golden}$ from ground truth annotations (for each sentence, there is only a few keywords, we assign zero to other non-keyword tokens). Then we are able to compute MSE score between the two aforementioned matrices: $MSE(A_{token\_label},A_{golden})$. We also reconstruct the token-label matrix from the BERT+single model as a comparison. RoBERTa has an MSE score of 0.0901 and RoBERTa-LiGCN has 0.0020. RoBERTa-LiGCN has a significant lower MSE score compared with RoBERTa. 
The T-test between the two models based on the predictions is 0.016, showing a significant difference. Since other datasets do not contain token-level annotations, so we fail to conduct quantitative analysis on them.

\textbf{Highlighted Tokens}
Additionally, in Table \ref{tab:highlight}, we show a case study selected from AAPD. We keep the top tokens highlighted only. This article is correctly classified as logic in computer science(cs.lo), programming languages (cs.pl) and software engineering (cs.se), marked by different colors. One can notice that the highlighted tokens are closely related to the class fields: \textit{object-oriented software} and \textit{Object Programs} are associated with cs.se; \textit{reference expressions} is associated with cs.pl; \textit{describing program semantics} is associated with cs.lo. Note that we omit highlighting of tokens that may appear in more than two classes for simplicity.


\definecolor{color1}{RGB}{247,194,183}
\definecolor{color2}{RGB}{187,230,190}
\definecolor{color3}{RGB}{187,230,222}
\renewcommand{\tabularxcolumn}[1]{>{\raggedright\arraybackslash\setlength{\parskip}{.5em}}p{#1}}

\begin{table*}[t]
	\centering
	\scriptsize
	\begin{tabularx}{\textwidth}{X}
	\toprule 
	Verifying properties of \colorbox{color1}{object-oriented software} requires a method for handling references in a simple and intuitive way, closely related to how \colorbox{color1}{O-O programmers} reason about their programs. The method presented here, a Calculus of \colorbox{color1}{Object Programs}, combines four components: \colorbox{color3}{compositional logic}, a framework for \colorbox{color3}{describing program semantics} and proving program properties; negative variables to address the specifics of \colorbox{color1}{O-O programming}, in particular \colorbox{color2}{qualified} \colorbox{color2}{calls};the alias calculus, which determines whether \colorbox{color2}{reference expressions} can ever have the same value...\\ \midrule[0.3pt]
	\textbf{Classes}: \colorbox{color1}{software engineering (cs.se)}, \colorbox{color2}{programming languages (cs.pl)}, \colorbox{color3}{logic in computer science (cs.lo)} 
		\\ 
		\bottomrule
	\end{tabularx}
	\caption{Highlighting tokens: two random paper abstracts in AAPD \cite{meyer1999towards}. Dark color means a higher correlation between token and classes.}
    \label{tab:highlight}
\end{table*}

\subsection{Label Correlations}

As we model class labels as nodes in the graph, we can then investigate if and how the learned label node representations are meaningful.
After training LiGCN, we take the label node representations of the last LiGCN layer and calculate cosine similarity between each label pair. We assume that the meaningful representations of a label pair should have a small angle in the latent space (i.e. their cosine similarity tends to the value of 1) if they have a positive correlation, and a large angle if they have a negative correlation.
We also investigate label correlations by looking at the model predictions. We collect model predictions in the test set and each label is represented as a binary vector with the dimension equal to the size of test set, and then calculate Pearson correlation between each label pair. Similarly, if Pearson correlation value tends to be 1, then it means a positive relationship; if the value tends to be -1, then it means a negative relationship.

\begin{figure}[t]
    \centering 
\begin{subfigure}{0.5\textwidth}
  \includegraphics[width=\linewidth]{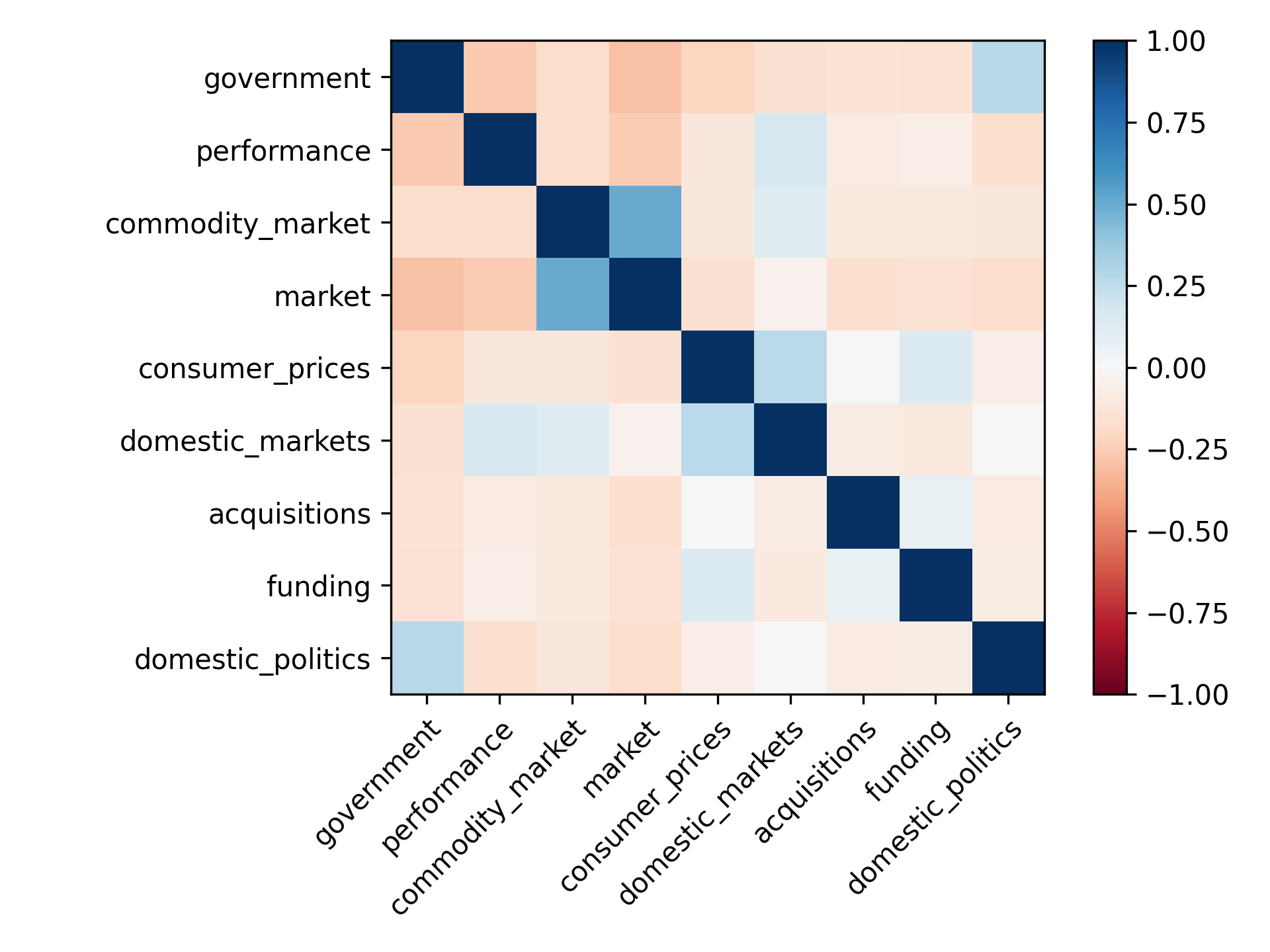}
  \caption{Pearson correlation.}

  \label{fig:rcv1_pearson}
\end{subfigure}\hfil 

\begin{subfigure}{0.5\textwidth}
  \includegraphics[width=\linewidth]{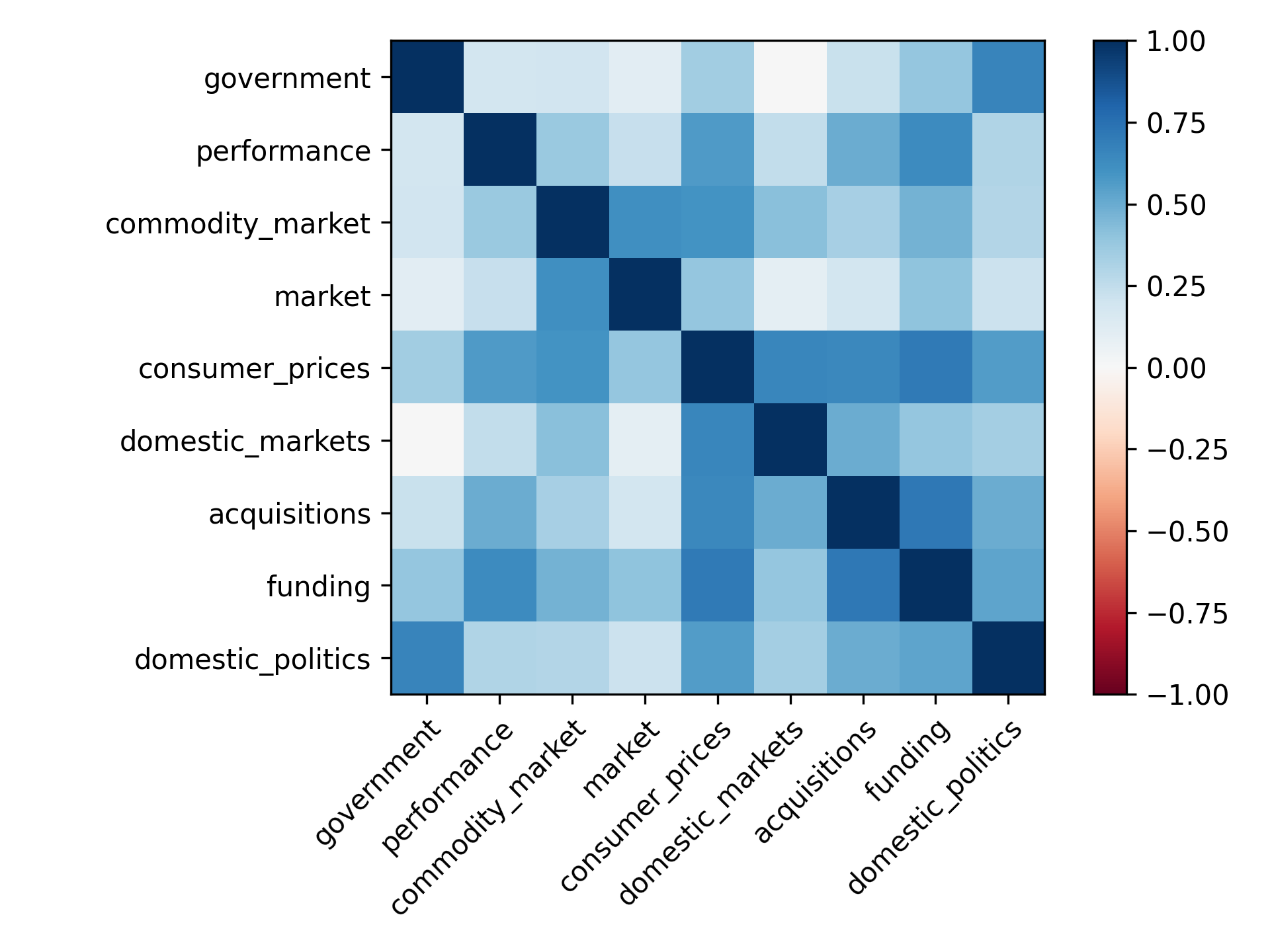}
  \caption{Label cosine similarity.}
  \label{fig:rcv1_cos}
\end{subfigure}\
  \vspace{-3mm}
\caption{Visualization of selected topics on RCV1. }
\label{fig:rcv1_ana}
\end{figure}

\textbf{Selected News Topics from RCV1} 
In RCV1 we select 9 topics to plot heatmaps randomly:
\textit{government, financial performance, commodity market, consumer prices, domestic markets, acquisitions, funding} and \textit{domestic politics}. Due to the limited space, we only show Pearson correlations and cosine similarities between each pair of our LiGCN model with the best performance in Figure \ref{fig:rcv1_ana}. For the Pearson correlation, we could notice that the model captures strong positive relationships between the following pairs: \textit{commodity market} and \textit{market}, \textit{government} and \textit{domestic politics}, \textit{consumer prices} and \textit{domestic markets}. These relationships are consistent with our real life, i.e., government news and domestic political news are very similar. We see a similar trend in the heatmap of cosine similarity for the mentioned label pairs. And in this way, more positive relations are found than negative ones, for example, negative correlation between \textit{acquisitions} and \textit{consumer prices}.

\section{Conclusion}

In this work, we propose a label-interpretable graph model, LiGCN, to solve the MLTC problem as a link prediction task. Our model is able to provide token-level explanation for the classification and therefore enjoys better label interpretability. Experiments on four public datasets show that our model achieved competitive scores. In the future, we will experiment with more complex graph encoders, extend this idea to single-label and extreme multi-label classification tasks \cite{neural-attention}.

\bibliography{anthology,custom}
\bibliographystyle{acl_natbib}

\end{document}